\documentclass{article} % For LaTeX2e

\PassOptionsToPackage{numbers, compress}{natbib}
\usepackage[preprint]{neurips2023/neurips_2023}
\usepackage{algorithm}
\usepackage{algpseudocode}

\usepackage[utf8]{inputenc} % allow utf-8 input
\usepackage[T1]{fontenc}    % use 8-bit T1 fonts
\usepackage{hyperref}       % hyperlinks
\usepackage{url}            % simple URL typesetting
\usepackage{booktabs}       % professional-quality tables
\usepackage{amsfonts}       % blackboard math symbols
\usepackage{nicefrac}       % compact symbols for 1/2, etc.
\usepackage{microtype}      % microtypography
\usepackage{xcolor}         % colors
\usepackage{mathtools}
\usepackage{enumerate}
\usepackage{caption}
\usepackage{amsthm}

\usepackage{subcaption}
\usepackage{graphicx}

\usepackage{natbib}

\usepackage{amsfonts,amsmath,amssymb}
\usepackage{bbm}
\usepackage{xspace}

\usepackage{multicol}
\usepackage{enumitem}

\providecommand{\mc}[1]{\mathcal{#1}}

\usepackage{hyperref}
\usepackage{url}

\newcounter{TesCounter}

\title{Comparing Foundation Models using Data Kernels}

\author{
Brandon Duderstadt\thanks{denotes equal contribution. Order of authors decided by a single coin flip.}\\
Nomic AI \\
\texttt{brandon@nomic.ai}
\And
Hayden~S.~Helm$^{*}$\\
Nomic AI \\
\texttt{hayden@nomic.ai}
\And
Carey~E.~Priebe \\
Johns Hopkins University \\
\texttt{cep@jhu.edu}
}

\begin{document}
\maketitle

\begin{abstract}
    Recent advances in self-supervised learning and neural network scaling have enabled the creation of large models, known as foundation models,  which can be easily adapted to a wide range of downstream tasks.
    The current paradigm for comparing foundation models involves evaluating them with aggregate metrics on various benchmark datasets.
    This method of model comparison is heavily dependent on the chosen evaluation metric, which makes it unsuitable for situations where the ideal metric is either not obvious or unavailable.
    In this work, we present a methodology for directly comparing the embedding space geometry of foundation models, which facilitates model comparison without the need for an explicit evaluation metric.
    Our methodology is grounded in random graph theory and enables valid hypothesis testing of embedding similarity on a per-datum basis.
    Further, we demonstrate how our methodology can be extended to facilitate population level model comparison.
    In particular, we show how our framework can induce a manifold of models equipped with a distance function that correlates strongly with several downstream metrics.
    We remark on the utility of this population level model comparison as a first step towards a taxonomic science of foundation models.
\end{abstract}

\section{Introduction}

In 2019, Devlin et al. introduced BERT \citep{bert}, a neural language model trained on massive amounts of unlabeled data which produces general purpose representations of language called embeddings.
BERT's embeddings can be used to dramatically reduce the amount of data and compute required to train a model on a downstream task.
BERT was the first of a class of models that have come to be known as foundation models, or large models trained in a self-supervised fashion which can be readily adapted to downstream tasks.
Recently, advances in language model scaling \citep{lm_few_shot}, prompt design \citep{chain_of_thought_prompting}, and modality fusion \citep{clip} have lead to the rapid and near ubiquitous adoption of foundation models across industry and academia alike \citep{foundation_opportunity_and_risks}.

Principled evaluation methodologies for foundation models have not kept pace with this mass adoption.
In particular, the most extensive attempts to characterize and evaluate large language models \citep{helm_not_hayden, mteb} involve benchmarking their performance on a wide variety of datasets using a set of aggregate performance metrics.
Unfortunately, this method of model comparison is unsuitable in situations where the ideal metric is either not obvious nor unavailable.
Further, even when an evaluation metric is known, it may provide an incomplete characterization of model performance.
For example, comparing models based on aggregate perplexity alone is not sufficient for understanding if they are disproportionately underperformant on a particular subgroup of data \cite{stochpar}.

An ideal model comparison methodology would surface exactly the set of data that differs between two models.
This would enable practitioners to \textit{discover} systematic differences between the models being compared, without having to define a metric which captures those differences a priori.
Towards this end, we propose a framework for directly comparing the geometry of the embedding spaces learned by different models on a per-datum basis.
The framework enables, for the first time, a comparison of the representations learned by foundation models that is agnostic to any particular performance metric.
We test our framework using a controlled training data ablation experiment, and find that it is able to surface changes in the representations of documents corresponding to the ablated class.

Due to the already large and continually increasing size of the foundation model design space, an ideal model comparison would also facilitate population level model comparison.
Accordingly, we demonstrate how to extend our framework to enable multi-model comparison by inducing a manifold of models.
We find that the distance between two models on this manifold is correlated strongly with the similarity of their performance on downstream tasks.
In future work, we aim to scale our population level comparison methodology to induce an empirical manifold of models.
Such a manifold would allow practitioners to frame model evaluation as a taxonomic problem, and generalize findings about the performance of a particular model to the performance of a family of models.

\textbf{Notation.}
We use lower case letters for vectors, upper case letters for matrices, $ \|\cdot\|_{2} $ to denote the Euclidean norm on vectors, and $ \|\cdot\|_{S} $ for the spectral norm on matrices. We use $ \hat{t} $ to denote an estimate of the population parameter $ t $.

\section{Modeling Embedding Space with the Data Kernel}
 \subsection{Defining the Data Kernel}
Before we can compare the embedding spaces of multiple foundation models, we require a model of the embedding space of a single foundation model.
%In our context we are interested in $ f $ as an embedding function from the model's input space $ \mc{X} $ (e.g., the space of text, images, etc.) to an embedding space $ \mc{Y} $. For convenience, then, we simply let $ f: \mc{X} \to \mc{Y} $ and assume that $ \mc{Y} = \mathbb{R}^{d} $ is a vector space.
Consider a foundation model $f: \mathcal X \to \mathcal Y$, where $\mathcal X$ is the model's input space (e.g., the space of text, images, etc.), and $\mathcal Y = \mathbb R^d$ is the model's embedding space (e.g., pooled word embeddings, image embeddings, etc.).
% For notational convenience, we describe our methodolgy in the context of vector embedding spaces $\mathcal Y = \mathbb R^d$.
For $N$ input examples $x_1, x_2, ..., x_N \in \mathcal X$, we compute the embeddings of these examples $f(x_1)=y_1, f(x_2)=y_2, ..., f(x_N)=y_N \in \mathbb R ^{d}$.
Let $X$ be the set $\{x_1, x_2, ..., x_N\}$ and $Y $ be the matrix whose rows are $y_1, y_2, ..., y_N$.
We define the \textit{data kernel} $A$ of $f$ on $X$ as
\begin{equation}
\label{eq:data-kernel}
    A = \text{TOP}_k(f(X)f(X)^{\top}) = \text{TOP}_{k}(YY^{\top})
\end{equation}
where $\text{TOP}_k: \mathbb R^{N,N} \to \{0, 1\}^{N,N}$ is a function applied to every cell in its input matrix that returns $1$ if that cell is in the top k values of its row (and not on the diagonal) and $0$ otherwise. In other words, $ \text{TOP}_{k} $ returns the hollow adjacency matrix of the $k$-nearest neighbor graph of its argument. 

We interpret the data kernel as the adjacency matrix of a graph where each node is an element in our input data, and an edge exists between nodes $i$ and $j$ if $j$ is one of $i$'s $k$ nearest neighbors.
Critically, the notion of distance used to define proximity when constructing this graph is induced by the foundation model $f$. 

Modeling data geometry with a neighborhood graph is an established technique with known desirable properties in the dimensionality reduction literature. For example, 
it is known that that the shortest paths on the weighted $k$ neighbor graph approximate the geodesic distances between datapoints under the metric induced by the model. \cite{isomap} 
%McInnes et al. 
The popular UMAP algoroithm \cite{umap} utilize the weighted $k$ neighbor graph to compute visualizations of high dimensional manifolds.
The unweighted $k$ neighbor graph, as described in in Eq.\ \eqref{eq:data-kernel}, retains many desirable limiting statistical properties, such as consistent estimation of the true-but-unknown underlying manifold density and metric \citep{lux_unweighted,consistent_unweighted}.
These properties motivate our use of the data kernel as a measurement of the geometry induced on a set of data by a particular model.

% \iffalse
% Figure \ref{fig:neighborhood_change} illustrates the effect of a particular noise model on a given data kernel,
% demonstrating that the position of a document in the visualization is sensitive to changes in the document's neighborhood structure.
%Figure \ref{fig:neighborhood_change} shows a set of visualizations of a data kernel with various degrees of corruption. 
%We note that the position of documents in the visualization is sensitive to changes in the document's neighborhood structure.
% This fact will form the basis of our qualitative comparison methodology.
% \fi

\subsection{Modeling Data Kernels as RDPGs}
\label{sec:dk_rdpg}
The Random Dot Product Graph (RDPG) \citep{Young2007RandomDP} is a flexible model for undirected, hollow, symmetric random graphs that has been applied successfully in a variety of domains \citep{larson2021dynamic, chen2022mental, winding2023connectome}. 
RDPGs posit that each node in a graph has an associated latent position $z$, and that the probability of an edge existing between nodes $i$ and $j$ in any realization of the RDPG is exactly the dot product of their corresponding latent positions $\langle z_i, z_j \rangle$.
%We model the data kernel as an RDPG because due to its successful application in a variety of domains \citep{larson2021dynamic, chen2022mental, doi:10.1126/science.add9330} and its corresponding theoretical foundations.
We write $ A \sim RDPG(ZZ^{\top})$ to refer to a graph $ A $ sampled from an RDPG with latent position matrix $ Z $. 

For analysis related to a single data kernel, methods developed under the RDPG assumption can rely on the consistent estimation of the latent positions via the adjacency spectral embedding \citep{avanti_rdpg}.
The consistency result holds up to the non-identifiability of an orthogonal transformation of the true-but-unknown latent positions, since $ \langle z ,z' \rangle = \langle Rz, Rz' \rangle $ for any orthogonal matrix $ R $.
%This nuance can cause issues when na\"ively comparing the estimated latent positions from two separate instances from the same RDPG model since the theorems only guarantee that there exists an orthogonal $ R $ such that the estimates $ \hat{z}^{(m)} $ and $ R\hat{z}^{(m')} $ from embedding the data kernels from foundation models $m $ and $m'$ are close.
%While there are efficient estimates of $ R $, such as Procrustes analysis \citep{schonemann1966generalized}, the estimation error can have serious impact on downstream inference. 

In the context of modeling and comparing the embedding spaces of foundation models using the RDPGs, the orthogonal non-identifiability of the parameters of the RDPG is quite natural.
For example, foundation models which employ cosine similarity to measure embedding proximity such as CLIP \citep{clip} and SBERT \citep{sbert} will give rise to identical data kernels under orthogonal transformation of their embedding spaces.
Vertex-level inference methods designed for observed graphs assumed to be an RDPG will thus be appropriate for datum-level analysis in the embedding spaces of the foundation models.

\subsection{Jointly Embedding Data Kernels}

Multi-graph analysis under the RDPG avoids the estimation of the unknown orthogonal matrix $ R $ by utilizing joint embedding techniques such as the omnibus embedding \citep{priebe2013manifold}. 
% With this data-kernels-as-graphs perspective, we can leverage existing multi-graph embedding techniques, such as the omnibus embedding \citep{priebe2013manifold, omni} to jointly embed and compare multiple data kernels.
In particular, let $ A^{(m)} $ and $ A^{(m')} $ be the symmetrized data kernels of a fixed input dataset $ X \in \mathcal{X}^{N} $ embedded by foundation models $f_{m} $ and $ f_{m'} $, respectively, and let $O_{m,m'} $ be the omnibus matrix defined as follows:
\begin{equation*}
O_{m,m'} = \mathcal{O}(A^{(m)}, A^{(m')}) =       \begin{bmatrix}
    A^{(m)} && \frac{A^{(m)} +A^{(m')}}{2} 
    \\
    \frac{A^{(m)} +A^{(m')}}{2} && A^{(m')} 
    \\
\end{bmatrix}.
\end{equation*}
We define $\hat {Z}^{(m)}_{m, m'}$ as the first $ N $ rows of the adjacency spectral embedding of $ O_{m,m'} $ and $ \hat{Z}^{(m')}_{m, m'}$ as the remaining $ N $ rows. When $ A^{(m)} $ and $ A^{(m')} $ are instances of the same RDPG model, both $ \hat{Z}^{(m)}_{m, m'} $ and $ \hat{Z}^{(m')}_{m, m'}$ are consistent estimates for the true-but-unknown underlying latent positions $ Z $ \citep{omni}. 
As a result, if two data kernels are drawn from the same underlying RDPG, the estimates of $ z_{i} $ after the omnibus embedding are such that:   
% be the estimate of the latent position for datum $i$ under the adjacency spectral embedding of $O$ for model $f_m$.
% Limit theory established in \cite{omni} states that, if $A^{(1)}$ and $A^{(2)}$ are realizations of RDPGs whose respective latent positions are equivalent up to orthogonal transformation:
\begin{equation}
\label{eq:consistency}
    \lim_{N \to \infty} \| \hat{z}_{i}^{(m)} - \hat{z}_{i}^{(m')} \|_{2} = 0 \text{ } \forall i.
\end{equation} 
% Furthermore, the estimates $\hat {Z}_i$ are asymptotically normal.
This convergence property is the key to our ability to compare model representation spaces, which we develop as a hypothesis test in the next section.
Intuitively, we can think of the omnibus embedding as a theoretically justified way to align different foundation models' embedding spaces for subsequent comparison.
%If two foundation models have embedding spaces that are the identical up to to orthogonal transformation, they will give rise to identical data kernels.
%We thus estimate these \textit{aligned} latent positions using the adjacency spectral embedding of the omnibus matrix. 

% These aligned estimates are not only asymptotically normal, but also converge to eachother as the size of the input dataset goes to infinity.

\section{Datum Level Hypothesis Testing}
\label{sec:comparing_data_kernels}
\subsection{A Bootstrap Hypothesis Test}

\begin{figure}[t]
    \centering
    \includegraphics[width=\textwidth]{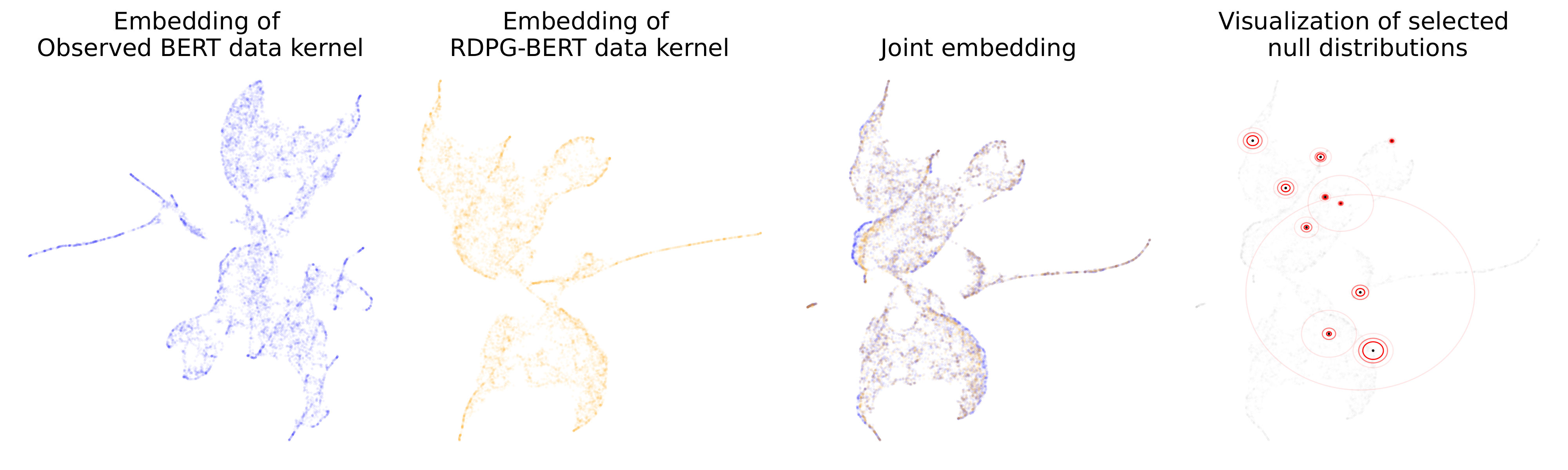}
    \caption{The UMAP projections of the adjacency spectral embeddings of the BERT data kernel (left), a data kernel sampled from RDPG-BERT (center-left),  and the joint embeddings (center-right) for a random subset of $10,000$ English Wikipedia articles. Each article is represented once in the left and center-left figures and twice in the center-right figure. The right panel shows the joint embeddings corresponding to the BERT data kernel with a random set of 10 articles emphasized. The concentric circles around the emphasized articles have radii equal to the 68th, 90th, and 99th percentiles of the bootstrap null distribution described in Section \ref{sec:hypothesis-testing}.}
    \label{fig:bert-rdpg}
\end{figure}

\begin{algorithm}[t!]
\caption{Bootstrap Hypothesis Test}\label{alg:bootstrap}
\begin{algorithmic}
\Require $X, f_1, f_2, k, n_{bootstrap}$
\State $A^{(1)} \gets \text{TOP}_k(f_1(X)f_{1}(X)^{\top})$
\State $A^{(2)} \gets \text{TOP}_k(f_2(X)f_{2}(X)^{\top})$
\State $\hat{Z}^{(1)} \gets \text{ASE}(A^{(1)})$ \Comment $\hat{Z}^{(1)} \in \mathbb{R}^{N,d}$ 
\State $T_{null} \gets []$
\For{$b \in \{1, 2, ..., n_{bootstrap}\}$}
    \State $A^{(b)} \sim RDPG(\hat Z^{(1)})$
    \State $O_{1,b} \gets \mathcal O(A^{(1)}, A^{(b)})$
    \State $\hat Z^{(1)}_{1, b}, \hat Z^{(b)}_{1, b} \gets \text{ASE}(O_{1,b})$
    \State $T_{null}\text{.append}(\|\hat{Z}^{(1)}_{1,b}-\hat{Z}^{(b)}_{1, b}\|_2)$ \Comment The norm is applied across the $d$ axis.
\EndFor
\State $O_{1,2} \gets \mathcal O(A^{(1)}, A^{(2)})$
\State $\hat Z^{(1)}_{1,2}, \hat Z^{(2)}_{1,2} \gets \text{ASE}(O_{1,2})$
\State $T \gets \|\hat{Z}_{1,2}^{(1)}-\hat{Z}_{1,2}^{(2)}\|_2$
\State $P \gets 1 - \text{Percentile}(T, T_{null})$ \Comment $P \in [0, 1]^N$ is vector of $p$ values - one for each document
\State \Return $P$
\end{algorithmic}
\end{algorithm}

\label{sec:hypothesis-testing}

Under the null hypothesis that the two data kernels are realizations of the same underlying RDPG, the limit theory outlined in \cite{omni} and briefly introduced in Eq.\ \eqref{eq:consistency} suggest a statistical hypothesis testing procedure for the embeddings induced by foundation models that is sensitive to changes in representations on a per-datum basis. 
In particular, if the two data kernels represent document $ i $ similarly then the distance $ d_{i} = \|\hat{z}_{i}^{(m)} - \hat{z}_{i}^{(m')}\|_{2} $ should be ``small'' for large enough $ N $. Conversely, if $ d_{i} $ is ``too large'' we may reconsider that the two foundation models represent document $ i $ differently. 

We formalize this statement via the following hypothesis test for document $ i $:
\begin{equation}
\label{eq:hypothesis-test}
H_0: z_i^{(m)} = z_i^{(m')} \quad \text{versus} \quad H_a: z_i^{(m)} \neq z_i^{(m')}.
\end{equation}
The hypothesis test requires defining a distance such that we can control for the Type 1 error and still reject the null when appropriate. 
To this end we propose estimating the null distribution of the distance $ d_{i} $ via a RDPG bootstrap.
%We prefer this approach, as opposed to leveraging the asymptotic normality of the estimated latent positions, due to its flexibility. For example, the bootstrap procedure we describe can be extended to transformations of the estimated latent positions such as UMAP and to different distances thereof. 
Algorithm \ref{alg:bootstrap} describes the bootstrap procedure for generating the null distribution of the $ d_{i} $ for each document in $ X $.
From the null distribution for document $ i $ and the observed distance $ d_{i} $ we obtain a $p$-value by subtracting the percentile at which $ d_{i} $ falls in the null distribution from 1 and reject $ H_{0} $ for $p$-values smaller than a pre-specified threshold. 
% Alternatively, the p-value can be used as a normalized distance to understand how much the representation of a document has changed across models relative to how much the representation changes under the null distribution.
%Figure \ref{fig:bert-rdpg} shows the UMAP \citep{umap} projections of the adjacency spectral embeddings $ \hat{Z} $ of the BERT data kernel (left), a data kernel sampled from the RDPG model parameterized by $ \hat{Z}\hat{Z}^{\top} $ (center-left), and the joint embedding of the corresponding omnibus matrix (center-right) of $ 10,000 $ random English Wikipedia articles. 
% In this experiment, and in the experiments below, the document embedding we use from BERT to construct the data kernel is the L2 normalized mean pooling of a document's word embeddings. \cite{sbert}
% The embeddings in the first and second embedding spaces are not natively comparable. 
% The joint embedding enables comparisons of the data kernels in a low dimensional embedding space. 
% The right panel of Figure \ref{fig:bert-rdpg} shows the UMAP projections corresponding to the BERT data kernel after jointly embedding with a data kernel sampled from RDPG($ \hat{Z} \hat{Z}^{\top}$). 
% The concentric circles around bolded points in the embedding correspond to the 68th, 90th, and 99th percentiles of $ d_{i} $.

Figure \ref{fig:bert-rdpg} visualizes several steps of the bootstrap when applied to the mean-pooled word embeddings of a pretrained BERT model.
In particular, we visualize the bootstrap of a data kernel that BERT induces on $ 10,000 $ English Wikipedia articles. 
The left panel shows the UMAP \citep{umap} projections of $\hat Z ^{(1)}$, the latent positions estimated from BERT's data kernel.
The center-left panel shows the UMAP projections of the latent position estimates from of the bootstrap data kernels, $ASE(A^{(b)})$.
Note that the left and center visualizations exhibit landmark level similarities (e.g. the long peninsula), but are not aligned.
The center-right panel shows the UMAP projection of $O_{1, b}$, one of the omnibus latent position estimates.
Note that the center-right panel has aligned the previously unaligned landmarks.
This alignment allows us to build a null distribution which captures how far latent position estimates from $A^{(b)}$ typically are from $A^{(1)}$.
The percentiles of selected null distributions are visualized using concentric circles in the far right panel.

% \textcolor{red}{for everyone: right now we only expose this framework for $ d_{i} = \|z_{i} - z_{i}'\|_{2} $. what level of generalization is useful? (i think for inclusion of UMAP, for example, we want something like $\Delta(g(z_{i}), g(z_{i}')) $  where $ \Delta(\cdot, \cdot) $ is some distance measure on the image of $ g $?}
% \textcolor{red}{bstadt: i think we should go for just $ d_{i} = \|z_{i} - z_{i}'\|_{2} $ in this paper, given the time constraints}

% \textcolor{red}{for CEP: do we automatically have validity of the proposed hypothesis test i) for the ASE xhats and ii) the umap o ase xhats from Keith's bootstrap? or do we need to extend the existing theoretical results?}

% for  p values for this hypothesis test.
% To peform the bootstrap, we first infer the parameters of a RDPG from $A^{(1)}$.
% Then, we sample a realization of this RDPG, $A^{(b)}$.
% Next, we compute the distance between the latent positions for each node in $\text{ASE}(\mathcal O\big(A^{(1)}, A^{(b)})\big)$.
% Finally, use the null distribution of distances between latent positions to evaluate, on a nodewise basis, whether the latent positions between the graphs we wish to test are the same or different.

% The P value of this test is interpreted as an estimate of the answer to the question "what is the probability that the latent positions estimated for the given node would be this far apart, assuming that both $A^{(1)}$ and $A^{(2)}$ are drawn from the same RDPG?"

\begin{figure}[t!]
     \centering
     \includegraphics[width=\textwidth]{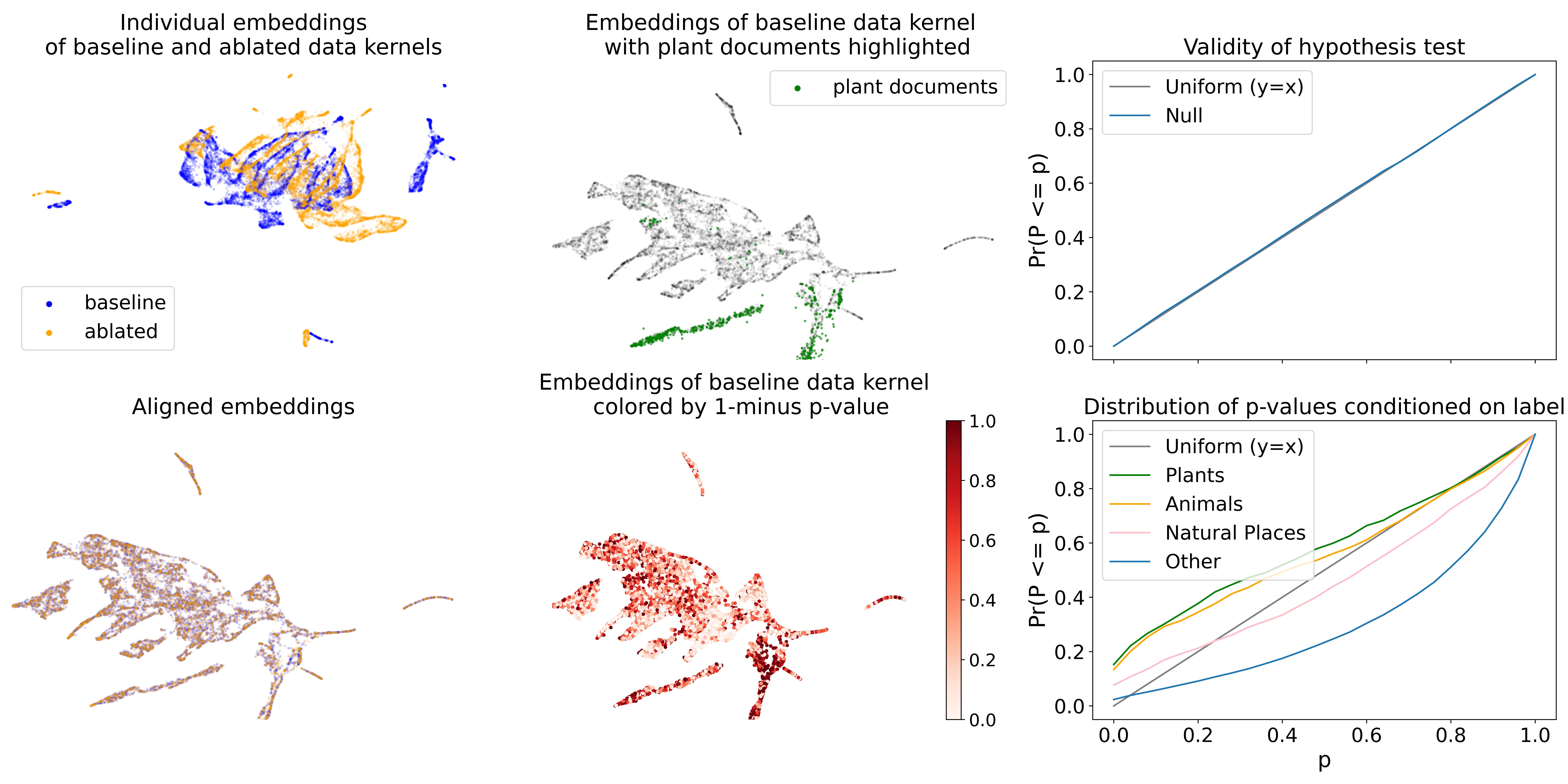}
     \caption{Individual embeddings (top left), aligned embeddings (bottom left), plant-highlighted (top center), comparison (bottom center), and bootstrapped hypothesis test (right) comparing the representations of language models trained on different corpora -- one baseline and one plant ablated.
     The data kernels under study are defined on a random subset of 10,000 documents from the DBPedia14 evaluation set. 
     }    
     \label{fig:carnivore}
 \end{figure}

\subsection{Training Data Ablation Study}
\label{subsec:data-ablation}
Under the emerging data-centric AI paradigm \cite{zha2023datacentric}, a model's behavior is largely determined by the data it is trained on.
If this is the case, we would expect interventions into a model's training data to affect its embedding space, and thus, its data kernel.
%We call this modification of a model's training data a data intervention.
% Common data interventions include new data collection and data curation.
%Oftentimes, aggregate statistics on a labeled dataset are the only way to measure the effectiveness of a data intervention.
%Unfortunately, labeled datasets are expensive to obtain and are often inaccurate quality measures due to pervasive label errors \cite{label_errors}.
%Furthermore, aggregate metrics may hide systematic errors present in a concentrated subset of the data distribution. Algorithm \ref{alg:bootstrap} can be leveraged to discover classes of data that are underrepresented in the training set. 

To test the effect of a training data intervention on the data kernel, we trained two randomly initialized BERT models -- a ``baseline" model and a ``plant-ablated'' model -- on two different versions of the DBPedia14 corpus \citep{dbpedia}.
The DBPedia14 training set consists of 40,000 examples of 14 different classes of documents such as Plants, Animals, Educational Institutions, etc. 
The baseline model is trained using masked language modeling (MLM) on all classes in the DBPedia14 train set.  
The plant-ablated model is trained using MLM  on all documents in the DBPedia14 training set except those belonging to the plant class.
We downsample the total size of the baseline model's train set uniformly across classes to ensure both models see the same number of tokens during training.
Both the plant-ablated model and the baseline model are trained for 3 epochs using the Adam optimizer \cite{adam} with a learning rate of $1e^{-4}$ and achieve comparable terminal losses. 
The data kernels we refer to below are defined on a random 10,000 article sample from the DBPedia14 evaluation set with $ k = 64 $.

Figure \ref{fig:carnivore} shows the result of applying the datum level hypothesis test to the baseline and plant-ablated models.
The top left panel of Figure \ref{fig:carnivore} shows the UMAP visualizations of the individual data kernels induced by each of the two models. 
Note that, while there are some structural similarities between the visualizations of the kernels, they are not directly comparable.
The bottom left panel of Figure \ref{fig:carnivore} shows the UMAP visualization of the aligned embeddings from both models. 
%There are no regions of the aligned embedding space that are populated exclusively by embeddings associated with only one of the plant-ablated model or the baseline model. 
%This indicates that our joint embedding procedure has effectively aligned the data kernels.
The top center panel of Figure \ref{fig:carnivore} highlights the regions of the aligned visualization that correspond to plant documents.
The bottom center panel of Figure \ref{fig:carnivore} shows colors the aligned baseline embeddings by 1-minus the datum-level $p$-value from Algorithm \ref{alg:bootstrap}. 
Larger values (represented here by hotter colors) indicate a higher probability that the baseline and plant ablated models' representations differ on a given document. 
Note that the two peninsulas consisting predominantly of plant articles appear ``hotter" than other regions.

The top right panel of Figure \ref{fig:carnivore} compares the $p$-values under the null distribution calculated using Algorithm \ref{alg:bootstrap} to the distribution of a uniform random variable. 
A hypothesis test is valid (i.e., has Type I error less than a pre-specified threshold) for all pre-specified thresholds if and only if the null distribution of $p$-values is uniform. 
In our case, the hypothesis test is valid. 
The bottom right panel compares the cumulative distribution of $p$-values for different classes of DBPedia14 documents. 
The $p$-values corresponding to plant documents have a left shifted distribution when compared to the $p$-values corresponding to the other classes. 
Additionally, the $p$-value distributions corresponding to animals and natural places, classes which are semantically related to plants, are also shifted left relative to the other classes.
This indicates that the baseline BERT used information contained in the articles about plants to inform its representations of articles about animals and natural places.

This ablation study highlights the ability of our datum-level test to identify systematic differences between models without a predefined metric.
Further, this procedure can be used to surface exactly the set of documents whose representations were affected by a data or model intervention.
These properties are useful in situations where an intervention systematically affects a class of data (e.g., data corresponding to a particular gender, race, or idea) that is neither explicitly modeled in the existing data ontology nor surfaced via a predefined metric.

\begin{algorithm}[t!]
\caption{Induce a model manifold}\label{alg:model-manifold}
\begin{algorithmic}
\Require $X, f_1, \hdots, f_{M}, k $
\For{$m \in \{1, 2, ..., M\}$}
    \State $ A^{(m)} \gets \text{TOP}_k(f_{m}(X)f_{m}(X)^T) $ \Comment{Construct data kernel}
\EndFor
\State $ D \gets M \times M \text{ matrix of zeros} $ 
\For{$m \in \{1, 2, ..., M\}$}
    \For{$m' \in \{1, 2, ..., M\}$}
        \State $ O_{m, m'} \gets \mathcal{O}(A^{(m)}, A^{(m')}) $
        \State $ \hat{Z}^{(m)}_{m, m'}, \hat{Z}^{(m')}_{m, m'} \gets ASE(O_{m, m'}) $
        \State $ D_{m,m'} = \| \hat{Z}^{(m)}_{m, m'} - \hat{Z}^{(m')}_{m, m'} \|_{S} $ \Comment{Distance between aligned embeddings}
    \EndFor
\EndFor
\State $ \hat{V} \gets \texttt{MultiDimensionalScaling}(D) $ \Comment{Euclidean representations of foundation models}
\State \Return $\hat{V}$
\end{algorithmic}
\end{algorithm}

\section{Multi-Model Comparison} 
\subsection{Inducing a Model Manifold}

The framework we have developed thus far enables datum level pairwise model comparison.
However, given the sheer size and variety of the model design space, it is desirable to be able to compare multiple models at once.
We can easily extend our pairwise comparison framework to allow for this multi-model comparison.

In section \ref{sec:comparing_data_kernels}, we defined a notion of distance between datum representations based on the Euclidean distance between their aligned latent position estimates. 
We can extend this notion of distance between datums to a notion of distance between models by considering the spectral norm of the difference between all of the aligned latent position estimates: $ \| \hat{Z}^{(m)}_{m, m'} - \hat{Z}^{(m')}_{m, m'} \|_{S} $. 
Under the assumption that $ A^{(m)} $ and $ A^{(m')} $ are realizations from the same underlying RDPG model, we have
\begin{equation}
    \lim_{N \to \infty} \frac{\| \hat{Z}^{(m)}_{m, m'} - \hat{Z}^{(m')}_{m, m'} \|_{S}}{\min(\|\hat{Z}^{(m)}_{m, m'}\|_{S},  \|\hat{Z}^{(m')}_{m, m'} \|_{S})} = 0, \label{eqn:spectral_norm}
\end{equation} where $ \hat{Z}^{(m)}_{m, m'} $ and $ \hat{Z}^{(m')}_{m, m'} $ are the corresponding aligned embeddings \citep{omni}.

Given foundation models $ f_{1}, \hdots, f_{M} $ and a document corpus $ X $, we can calculate the spectral norm of the differences between aligned latent position estimates of the data kernels for each pair of models with the understanding that if $ \| \hat{Z}^{(m)}_{m, m'} - \hat{Z}^{(m')}_{m, m'} \|_{S} < \| \hat{Z}^{(m)}_{m, m''} - \hat{Z}^{(m'')}_{m, m''}\|_{S} $ then $ f_{m} $ is in some sense ``closer" to $ f_{m'} $ than $ f_{m''} $ (with respect to $ X $). 
Further, we can infer the relative positions of these models on a \textit{model-manifold} from their pairwise distances using classical multi-dimensional scaling.
%In order to infer more complicated relationships given the pairwise distances of a collection of models, we can employ multi-dimensional scaling.

Classical multi-dimensional scaling \citep{torgerson1952multidimensional} recovers the unknown relative positions of a collection objects given a pairwise Euclidean distance matrix $ D $. 
In particular, for objects $ V^{(1)}, \hdots, V^{(M)} \in \mathbb{R}^{d} $ where $ D_{m,m'} = \|V^{(m)} - V^{(m')}\|_{2} $ is a pairwise distance matrix on the $ V^{(m)}$'s, classical multi-dimensional scaling will recover the relative positions of each $ V $ up to an orthogonal transformation.
Recovery of the relative positions of objects with multi-dimensional scaling in more general contexts is possible when the objects and dissimilarity used to construct $ D $ are Euclidean realizable \citep{borg2005modern}.

In our case, we assume that there is a true-but-unknown vector representation $ V^{(m)} \in \mc{V} $ of the the foundational model $ f_{m} $ in model space with respect to $ X $ and that $ \hat{Z}^{(m)} $ is a matrix-valued representation of $ V^{(m)} $. 
If we consider the space $ \mc{V} $ as a parameterization of the space of RDPGs, the dissimilarity matrix with entries $ D_{m, m'} = \| \hat{Z}^{(m)}_{m, m'} - \hat{Z}^{(m')}_{m, m'} \|_{S} $ is Euclidean realizable given some regularity conditions on $ \mc{V} $ \citep{athreya2022discovering}. 
The multi-dimensional scaling of $ D $ yields vectors $ \hat{V}^{(1)}, \hdots, \hat{V}^{(M)} \in \mathbb{R}^{d} $ that are Euclidean approximations of the foundation models $ f^{(1)}, \hdots, f^{(M)} $ relative positions in model space with respect to $ X $. 
We refer to the space spanned by $ \hat{V}^{(1)}, \hdots, \hat{V}^{(M)} $ as the ``model manifold" with respect to $ X $. 
This process -- with inputs of a collection of foundation models and a document corpus $ X $ -- is described in Algorithm \ref{alg:model-manifold}.

\subsection{The Manifold of Partially Ablated Models}
\begin{figure}[t]
    \centering
    \includegraphics[width=\linewidth]{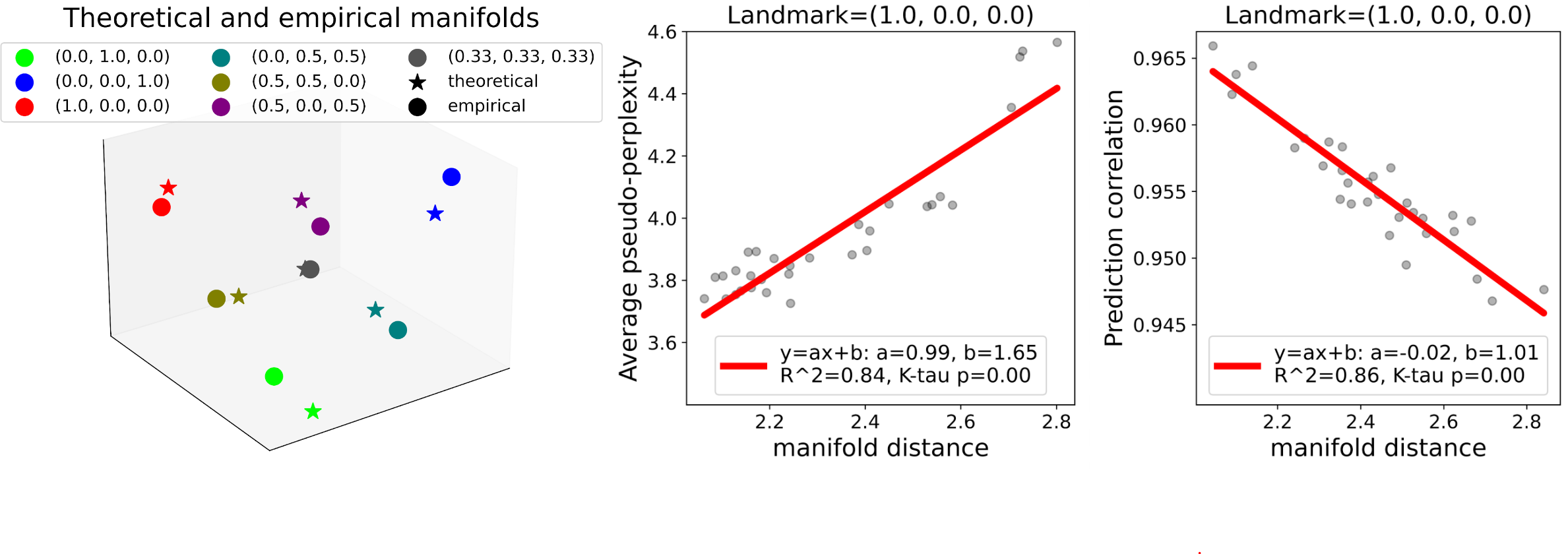} 
 %    \begin{subfigure}{\textwidth}
 %          \centering
 %          % include second image
 %          \includegraphics[width=\linewidth]{figures/Model-Manifold/3d-theoretical-versus-empirical.pdf}  
 %        % \caption{Visualization of the theoretical (stars) and empirical (dots) model manifold.}
 %        \label{subfig:manifolds}
 %    \end{subfigure}
 %    \begin{subfigure}{0.3\textwidth}
 %          \centering
 %          % include second image
 %          \includegraphics[width=\linewidth]{figures/Model-Manifold/prediction_correlation_versus_distance_to_landmark_1_0_0.pdf}
 %        % \caption{Inference similarity versus inverse manifold distance.}
 %        \label{subfig:correlation}
 %    \end{subfigure}
 %    \begin{subfigure}{0.3\textwidth}
 %          \centering
 %          % include second image
 %          \includegraphics[width=\linewidth]{figures/Model-Manifold/perplexity_versus_distance_to_landmark_1_0_0.pdf}
 %        % \caption{Inference similarity versus inverse manifold distance.}
 %        \label{subfig:perplexity}
 %    \end{subfigure}
    \caption{Theoretical (the 2-d simplex) and empirical manifolds induced via multi-dimensional scaling of low-dimensional representations of data kernels (left). The manifold distance correlates strongly with classifier similarity to a landmark model (right) and pseudo-perplexity on a language modeling task (center).}
    \label{fig:model-manifold}
\end{figure}

To investigate the properties of the model manifold, we build on the experiment described in Section \ref{subsec:data-ablation}. 
Instead of considering just a plant-ablated model and a baseline model, we now consider models that are trained on various ablated subsets of DBPedia14. 
We find that the relative positions of models on the manifold induced by our procedure are approximately parameterized by the relative concentrations of DBPedia14 classes in the model training sets.
Further, we find that the distance between models on the manifold correlates strongly with the difference in the models' performance on several downstream metrics.
This indicates that proximity on the model manifold may be used as an effective proxy measure for similarity in model performance.

%All of the models we trained had access to the 40,000 documents from each class of DBPedia except for the Plant, Artist, and Educational Institution classes. 
Again, we trained all models on ablated subsets of DBPedia14 using the masked language modeling objective for 3 epochs using the Adam optimizer and a learning rate of $1e-4$.
This time, each model had the training data corresponding to some combination of the Plant, Artist, and Educational Institution classes ablated.
Recall that each class in the DBPedia14 dataset has 40,000 total documents in it.
To determine the particular ablation profile for a model $ m $, we first selected an element of the two-dimensional simplex $ (p_{1}^{(m)}, p_{2}^{(m)}, p_{3}^{(m)}) \in \Delta^{2}$ (i.e., elements of $ \mathbb{R}^{3} $ whose elements are all non-negative and that sum to 1). 
The model then had access to $ p_{1}^{(m)} \cdot 40,000 $ documents from the Educational Institution class, $ p_{2}^{(m)} \cdot 40,000 $, documents from the Artist class, and $ p_{3}^{(m)} \cdot 40,000 $ documents from the Plant class. 
The documents from the remaining classes in the training data were then uniformly downsampled to ensure that all models had access to the same number of training tokens.
Hence, each model's training set is naturally parameterized by its corresponding element of the 2-d simplex $ (p^{(m)}_{1}, p^{(m)}_{2}, p^{(m)}_{3}) $.

The left panel in Figure \ref{fig:model-manifold} shows the theoretical manifold (the two-dimensional simplex which defines the class concentrations in each model's training set) and the empirical manifold found when applying Algorithm \ref{alg:model-manifold} with $k=64$ and aligning the empirical manifold with the theoretical manifold via the Procrustes algorithm. 
Here, models parameterized by $ (1, 0, 0), (0, 1, 0), (0, 0, 1), (0.5, 0.5, 0), (0, 0.5, 0.5), (0.5, 0, 0.5), $ and $ (0.33, 0.33, 0.33) $ are used to estimate the and visualize model manifold. 
The empirical manifold is visually similar to the theoretical manifold, and we interpret this as evidence that the models, through their data kernels, exist on a low dimensional structure that (in this case) is parameterized by the concentrations of classes in each model's training set. More succinctly, $ \hat{V}^{(m)} \approx (p^{(m)}_{1}, p^{(m)}_{2}, p^{(m)}_{3}) $ in this experiment. 

% Once all of the models were trained, we randomly selected 500 documents per class (7,000 total documents) and applied Algorithm \ref{alg:model-manifold} to obtain an empirical model manifold. 
%Since we assume that the data kernels are realizations from different RDPG models, the natural distance on the empirical model manifold is the Euclidean distance.
To understand how different downstream metrics correlate with distance on the recovered model manifold, we compute a metric relevant to document classification and a metric relevant to language modeling and regress them with respect to the manifold distance to a landmark model. For both tasks we select the model parameterized by $ (1, 0, 0) $ (i.e. the model trained on all of the Educational Institution articles, but none of the Artist or Plant Articles) as the landmark model.

For the classification task we consider a uniform subsample (500 documents per class) of the DBPedia14 evaluation set.
We train a linear support vector machine (SVM) on the mean-pooled embeddings from each model using a random 80\% of the selected evaluation data, and then infer the class membership of the remaining 20\%.
%For each model, we calculate the normalized frequency for which the class predict
ion for a document agrees with the prediction from the classifier trained on embeddings from the landmark model.
We then compute the correlation between the predictions of each model's SVM and the landmark model's SVM.
In the right panel of Figure \ref{fig:model-manifold}, we show that manifold distance between models is a strong indicator of their prediction correlation.
%We refer to this normalized frequency as the prediction correlation and report the average prediction correlation from 10 different random samples in Figure \ref{fig:model-manifold}. We expect models that were trained with more Educational Institution documents to have a higher prediction correlation with the landmark model.

For the language modeling task, we randomly sampled $ 1,000 $ documents in the Educational Institution class from the DBPedia14 evaluation set.
We compute the average pseudo-perplexity \citep{Salazar_2020} of each document in our evaluation set, and report the average pseudo-perplexity across all evaluation documents for each model in the center panel of figure \ref{fig:model-manifold}.
%We masked a token of a document and measured the pseudo-perplexity \citep{salazar2019masked} for the actual token. We repeat this process for each token in the document. The psuedo-perplexity of the document is the average pseudo-perplexity across the tokens. The pseudo-perplexity of the model, which we report, is the average pseudo-perplexity across all the evaluation documents.
We find that models that were trained with a higher concentration of Educational Institution documents have both a lower average pseudo-perplexity and are closer on the manifold 
We find that the concentration of Educational Institution documents in the training set correlates negatively with both average pseudo-perplexity and manifold distance to the landmark model

%The center and right panels of Figure \ref{fig:model-manifold} demonstrate that the distance on the model manifold strongly correlates with performance on downstream classification and language modeling metrics.
%For both tasks, the model manifold contains 32 models. 25 of the models are parameterized by random elements of the two-dimensional simplex.
%The remaining 7 are the models used to visualize the empirical manifold in the left panel of Figure \ref{fig:model-manifold}.
%The model corresponding to $ (1, 0, 0) $ (i.e., the model with access to only data from the Educational Institution class), is used as the landmark model.
%The manifold for the classification task (center) is with respect to a random stratified $7,000 = 14 \cdot 500 $ documents from the DPBedia14 evaluation set. 
% \textbf{TODO HHELM i dont know what this means. is the manifold for the classification task different than the one on the left? and the sentence the "manifold... is with respect to" doesn't seem to make grammatical sense. please be more precise here}
%The data kernel for each model is induced with $ k = 64 $.
%The manifold for the text generation task (right) is with respect to a random $ 1,000 $ documents from the Educational Institution class.
%The data kernel for each model is induced with $ k = 32 $. We use a larger $ k $ for the classification experiment because of the larger number of documents present in the evaluation set. 

The linear goodness-of-fit ($R^{2}$) for both the relationship between manifold distance and prediction correlations as well as the relationship between manifold distance and pseudo-perplexity are above 0.8, and the $p$-values corresponding to Kendall's tau rank correlation test are less than 0.01. 
We interpret these results as evidence that the manifold distance and metrics related to important downstream tasks -- classification and language modeling -- are strongly correlated.

% illustrates a potential use case of the model manifold for downstream inference. 
% For this experiment, we trained an additional 15 models that are parameterized by randomly selected elements of the two-dimensional simplex.
% We fit a line regressing the correction correlation to the inverse of the manifold distance for the models parameterized by the vertices of the two-dimensional simplex (or, the models with access to documents pertaining only to Plant, only to Educational Institution, and only to Artists). 
% The linear goodness-of-fit values ($ R^{2} $) are all above 0.60 and two of them are larger than 0.8. 
% We also report the p-value from the Kendall's tau rank correlation test for each set of points. While the tests across the figures are not independent, the p-value for each, up to two decimal places, is $ 0.00 $ and they give strong evidence that the rankings of the two metrics (inverse manifold distance and correctness correlation) are not independent.

\section{Limitations and Broader Impact}
\subsection{Limitations}
While we are excited about the potential of data kernel based foundation model comparison, there are several important limitations of our current work.

The most salient limitation is the scope of our experiments -- we have opted to evaluate the behavior of our comparison framework in several highly controlled situations.
The benefit of this choice is that we can ensure that the framework behaves appropriately under known conditions.
The limitation of this choice is that it provides limited evidence that the framework will generalize to less controlled, empirical situations.
We aim to address this question of generalization in subsequent work on the empirical model manifold.

An additional limitation of our work is the computational burden required for data kernel based comparison. 
Each iteration of our bootstrap procedure requires the computation of a large singular value decomposition as part of the adjacency spectral embedding. 
This presents a significant challenge to practitioners who wish to access our comparison methodology, but have a limited compute budget.
We encourage future work on more efficient bootstrap methods that can help to alleviate this burden, and aim to address it as part of our future research agenda.

\subsection{Broader Impact}
While we view the development of higher fidelity model evaluation procedures as generally positive from a broader impact point of view, there are a few important caveats to note.
As is the case for any evaluation, there is a risk that our framework fails to surface a systematic bias present in a model.
This risk is particularly present in our work, as a failures of our framework may lead practitioners to have a false sense of security regarding the fairness of their models.
We emphasize that model evaluation is a nuanced and complex question, and that the best approaches combine several model evaluation strategies to paint a maximally holistic picture of model behavior.

Moreover, the computationally intensive nature of the work has a direct environmental impact.
In particular, we estimate that this work produced at most 3 metric tons of carbon dioxide emissions, which is roughly the equivalent of burning 350 gallons of gasoline.
Nomic is committed to tracking and mitigating the environmental impact of its research, and has offset 3 metric tons of emissions for this project.

\section{Future Work}
In this work, we introduce a methodology to compare foundation models  by directly comparing their embedding space geometry. 
% a metric free methodology \tes{I will recommend reword metric-free to say subsequent inference/downstream evaluation free. My worry is that the method of comparing model embedding space geometry makes use of the Euclidean metric (like spectral norm), which sounds a bit contradicting to the metric free wording} 
% for comparing foundation models via their embedding space geometry.
We demonstrate how to apply this methodology to surface the populaton of data points whose representations differ between models.
We then extend this framework to multi-model comparison, and show that distance on the resulting model manifold correlates with common downstream performance metrics.
We envision several extensions of this work, some of which are outlined below.

\subsection{An Empirical Model Manifold}
Given the large and growing variety of foundation model variants, we are interested in using our procedure to build a manifold of empirical models. 
Technologies such as the Hugging Face Hub \citep{wolf2020huggingfaces} enable us to build model manifolds on large collections of models and datasets.
Having access to a large empirical model manifold will allow researchers to focus their analysis on models that exist in key locations on the manifold (e.g. centers of mass, branch points, etc.).
Additionally, having a notion of distance between empirical models will allow researchers to generalize their findings regarding particular models to \textit{families} of models that are proximal on the manifold.

\subsection{Model Selection}
The model manifold comes naturally equipped with a distance that captures information pertinent to relative performances of the models.
We can leverage this fact for a variety of model selection tasks.
In particular, the model manifold allows model selection to be framed as a search task over the model manifold itself. 
For example, users can select the best model for a particular task with far fewer evaluations than an exhaustive search by employing a locally sensitive optimizer on the model manifold.
Additionally, the task of selecting the best model for a task subject to a constraint (e.g., a compute constraint) given a known good model that does not fulfill that constraint is akin to projecting the known good model onto a constrained subset of the model manifold.

\subsection{Privacy Aware User Modeling}
The model manifold can be employed to study populations of users communicating on a platform without requiring direct access to the content of their communications.
One such procedure would involve fine tuning an edge language model to an individual user's communications, and relaying only the embeddings of the fine tuned model on a public corpus of interest back to the central server.
A model manifold where each model corresponds to a user could then be constructed and utilized for subsequent inference.

\subsection{Efficient RDPG Bootstrap}
One of the fundamental challenges of scaling up our work is the computationally intensive nature of our bootstrap procedure.
In particular, the singular value decomposition required for each iteration of our bootstrap presents a significant computational bottleneck.
There is a rich and active literature on efficient singular value decomposition algorithms, and we hope to evaluate the viability of ideas from this literature in future versions of our bootstrap.

\clearpage

% \input{text/introduction.tex}
% \label{introduction}

% \input{text/background.tex}
% \label{background}

% \input{text/methods.tex}
% \label{method}

\clearpage

\bibliographystyle{iclr2021/iclr2021_conference}
\bibliography{biblio}

\clearpage

% \label{appendix}

\end{document}